\documentclass[11pt]{article}

\usepackage[final]{acl}

\usepackage{times}
\usepackage{latexsym}
\usepackage{listings}

\usepackage[T1]{fontenc}

\usepackage[utf8]{inputenc}

\usepackage{multirow}
\usepackage{url}
\usepackage{microtype}

\usepackage{inconsolata}

\usepackage{graphicx}
\usepackage{algorithm}
\usepackage{algorithmic}
\usepackage{amsmath}
\usepackage{amssymb}
\usepackage{booktabs}
\usepackage{times}
\usepackage{helvet}
\usepackage{courier}
\usepackage{xcolor}
\usepackage{enumitem}
\usepackage[table]{xcolor}
\usepackage{subcaption}
\usepackage{hyperref}
\usepackage{url}
\usepackage{booktabs}

\usepackage{tcolorbox}
\tcbuselibrary{listings}
\tcbuselibrary{skins, breakable}

\title{Evo-PI: Aligning Medical Reasoning via Evolving Principle-Guided Supervision} 

\author{
  Xianda Zheng\textsuperscript{1} \quad
  Huan Gao\textsuperscript{6} \quad
  Meng-Fen Chiang\textsuperscript{4} \quad
  Michael Witbrock\textsuperscript{1} \\
  {\bf Kaiqi Zhao}\textsuperscript{\textbf{5}} \quad
  {\bf Shangyang Li}\textsuperscript{\textbf{2,3*}} \\
  \textsuperscript{1}School of Computer Science, University of Auckland\\
  \textsuperscript{2}Beijing University of Posts and Telecommunications \quad
  \textsuperscript{3}Renyixun Health Technology Co., Ltd. \\
  \textsuperscript{4} National Yang Ming Chiao Tung University \quad
  \textsuperscript{5}Harbin Institute of Technology (Shenzhen) \quad
  \textsuperscript{6}Sapiens \\
  \texttt{xzhe162@aucklanduni.ac.nz} \quad
  \texttt{nic\_lab@163.com} \\
}

\newcommand\blfootnote[1]{%
  \begingroup
  \renewcommand\thefootnote{}\footnote{#1}%
  \addtocounter{footnote}{-1}%
  \endgroup
}

\begin{document}
\maketitle

\blfootnote{$^*$Corresponding author.}

\begin{abstract}
Despite recent progress, the reasoning capabilities of large multimodal language models (MLLMs) remain fundamentally constrained by static supervision, where fixed prompts, rules, or reward models provide non-adaptive guidance throughout training. Such static signals are often sufficient to enforce output formats, but fail to shape the underlying reasoning process, leading to brittle generalization and performance saturation in complex decision-making tasks.
We propose \textbf{Evo-PI}, a principle-centric learning framework that treats reasoning principles as explicit, language-based supervision signals that can be generated, evaluated, and iteratively evolved. Instead of relying on fixed rewards, Evo-PI enables a co-evolutionary loop in which principles guide model reasoning, while model behaviors in turn refine the principles that supervise them. This dynamic alignment mechanism allows supervision to progressively adapt to the model’s reasoning deficiencies.
We instantiate Evo-PI in medical visual question answering as a high-stakes testbed requiring structured visual–textual reasoning. Across eight benchmarks and multiple model backbones, Evo-PI consistently improves reasoning accuracy, achieving gains of up to 24.6\%. Our results suggest that evolving principle-guided supervision offers a scalable and general paradigm for training expert-aligned reasoning in MLLMs.
Code is available at \url{https://github.com/zhengxianda/Evo_PI}.
\end{abstract}

\section{Introduction}
\label{sec:introduction}

Large multimodal language models (MLLMs) have demonstrated remarkable capabilities in integrating visual and textual information to perform complex reasoning~\cite{gemini2.5, GPT4, qwen2.5vl, HuatuoGPT-Vision, PathVLM-R1, MedVLM-R1, zhi2025medgr}. 
Despite their promise, current MLLMs remain fundamentally constrained by static supervision, where fixed prompts, rules, or reward models provide non-adaptive guidance throughout training~\cite{Med-R1, GRPO}. 
While such supervision can enforce output formats or improve final answer accuracy, it rarely shapes the underlying reasoning process, often leading models to learn brittle heuristics instead of robust, expert-like logic~\cite{Medical_LLM_1, Medical_LLM_2, Reward_Hacking,jing2025beyond}.

\begin{figure}[t] 
    \centering
    \includegraphics[width=1\linewidth]{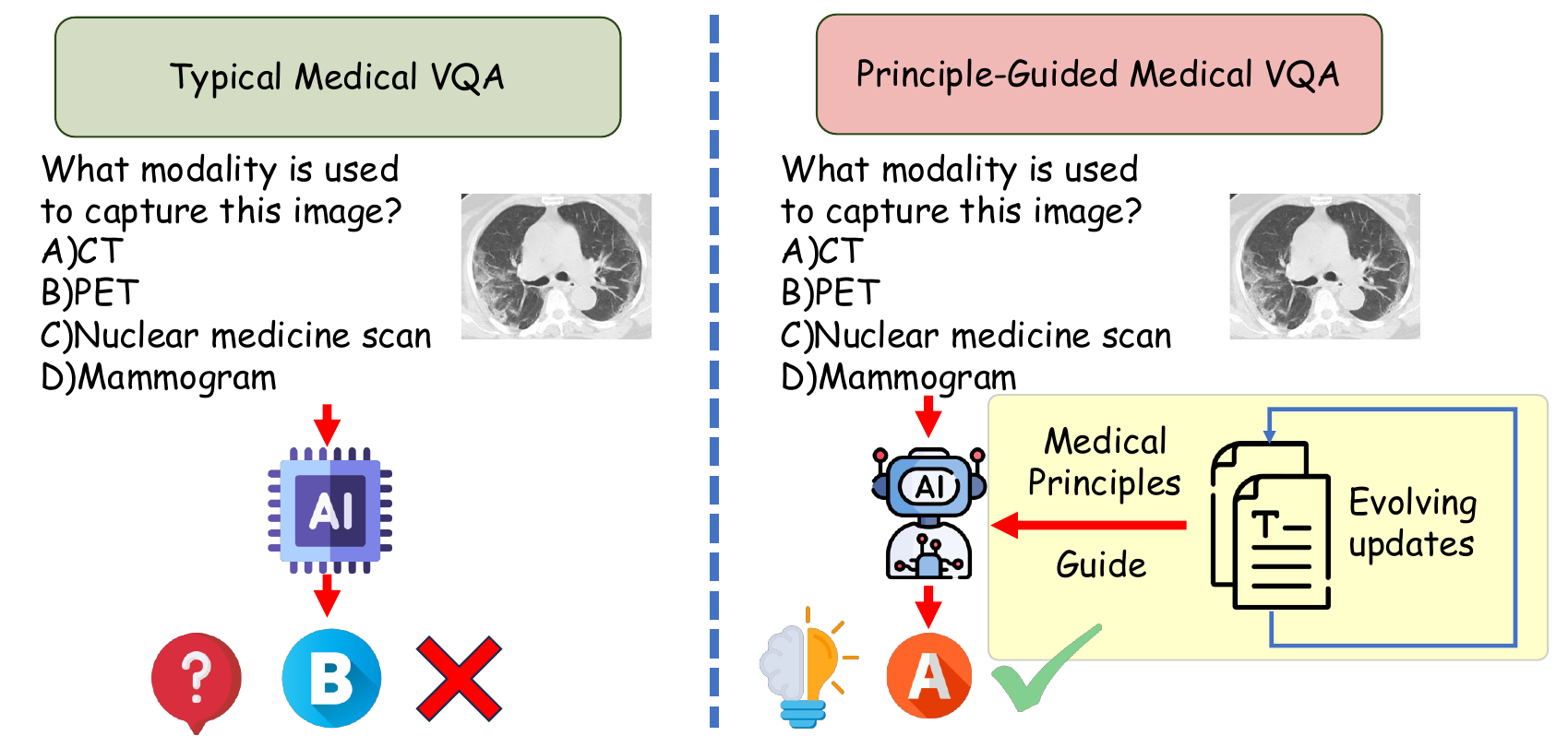} 
    \caption{
    Comparison between standard Medical VQA and our principle-guided framework.
    (Left) Without explicit guidance, the MLLM relies on superficial visual cues and fails in reasoning.
    (Right) With evolving medical principles as guidance, the model follows a structured, clinically aligned reasoning process and reaches the correct answer.
    }
    \label{illustration}
\end{figure}

Supervised fine-tuning (SFT) on annotated datasets has been the primary method for aligning MLLMs to specific reasoning tasks~\cite{HuatuoGPT-Vision, Med-R1, MedVLM-R1}. 
However, the acquisition of high-quality, expert-annotated data is labor-intensive and expensive, creating a bottleneck for scaling reasoning capabilities~\cite{Medical_LLM_1, Medical_LLM_2}. 
Reinforcement learning (RL) methods have been explored to address these limitations, optimizing models with scalar reward signals derived from accuracy, format compliance, or other heuristics~\cite{GRPO, DAPO, GSPO, Pass_k}. 
While these methods can improve generalization beyond SFT, they still act as ``fixed tutors'', providing limited feedback that enforces output correctness but does not adaptively shape intermediate reasoning~\cite{deepseek}. 
Consequently, models often plateau in performance, fail to generalize to complex cases, and remain prone to reward hacking~\cite{Reward_Hacking}.

Human learning provides a compelling analogy for overcoming these limitations. 
Experts do not rely solely on pass/fail evaluations; instead, they acquire abstract reasoning principles, apply them to diverse cases, and iteratively refine them based on experience~\cite{Medical_principles_1, Medical_principles_2}. 
This process enables transfer to unfamiliar or complex situations and supports robust, flexible decision-making. 
Inspired by this cognitive process, we hypothesize that \emph{evolving principles}, rather than static scalar rewards, should serve as primary supervision signals to guide MLLMs toward expert-level reasoning. Figure~\ref{illustration} illustrates the key insights of our idea.  

To operationalize this idea, we introduce \textbf{Evo-PI}, a framework for \textbf{Aligning Medical Reasoning via Evolving Principle-Guided Supervision}. 
Evo-PI establishes a co-evolutionary loop in which (i) a principle bank distills initial reasoning guidelines from curated examples, (ii) the backbone MLLM receives dense feedback on its adherence to these principles via a frozen judge model, and (iii) the principles themselves are iteratively refined based on the model’s evolving failures~\cite{MedVLM-R1, PathVLM-R1}. 
This dynamic alignment mechanism transforms static supervision into a continuous dialogue between model and principles, progressively narrowing the gap between the model’s reasoning and expert standards. 

We evaluate Evo-PI in the high-stakes domain of medical Visual Question Answering (VQA)~\cite{MLLM_Survey_med}, which requires integrating visual evidence with structured medical knowledge. 
Across eight benchmarks and multiple MLLM backbones, Evo-PI consistently improves reasoning accuracy, achieving gains of up to 24.6\% compared to models trained with static supervision. 
Qualitative analysis further demonstrates that Evo-PI produces more coherent and clinically valid reasoning traces, validating the effectiveness of language-based supervision in shaping intermediate reasoning rather than merely optimizing final answers.

In summary, our contributions are:
\begin{itemize}
    \item We propose \textbf{Evo-PI}, a principle-guided supervision framework that replaces static scalar rewards with evolving, language-based reasoning principles, mimicking the iterative learning process of human experts.
    \item We introduce a co-evolutionary mechanism where the reasoning model and its supervising principles mutually refine each other, mitigating reward hacking and enhancing reasoning robustness.
    \item We empirically demonstrate that this evolving supervision paradigm scales reasoning capabilities across diverse MLLM backbones and medical VQA testbeds, providing a generalizable path toward expert-aligned multimodal reasoning.
\end{itemize}

\begin{figure*}[t] 
    \centering
    \includegraphics[width=0.95\textwidth]{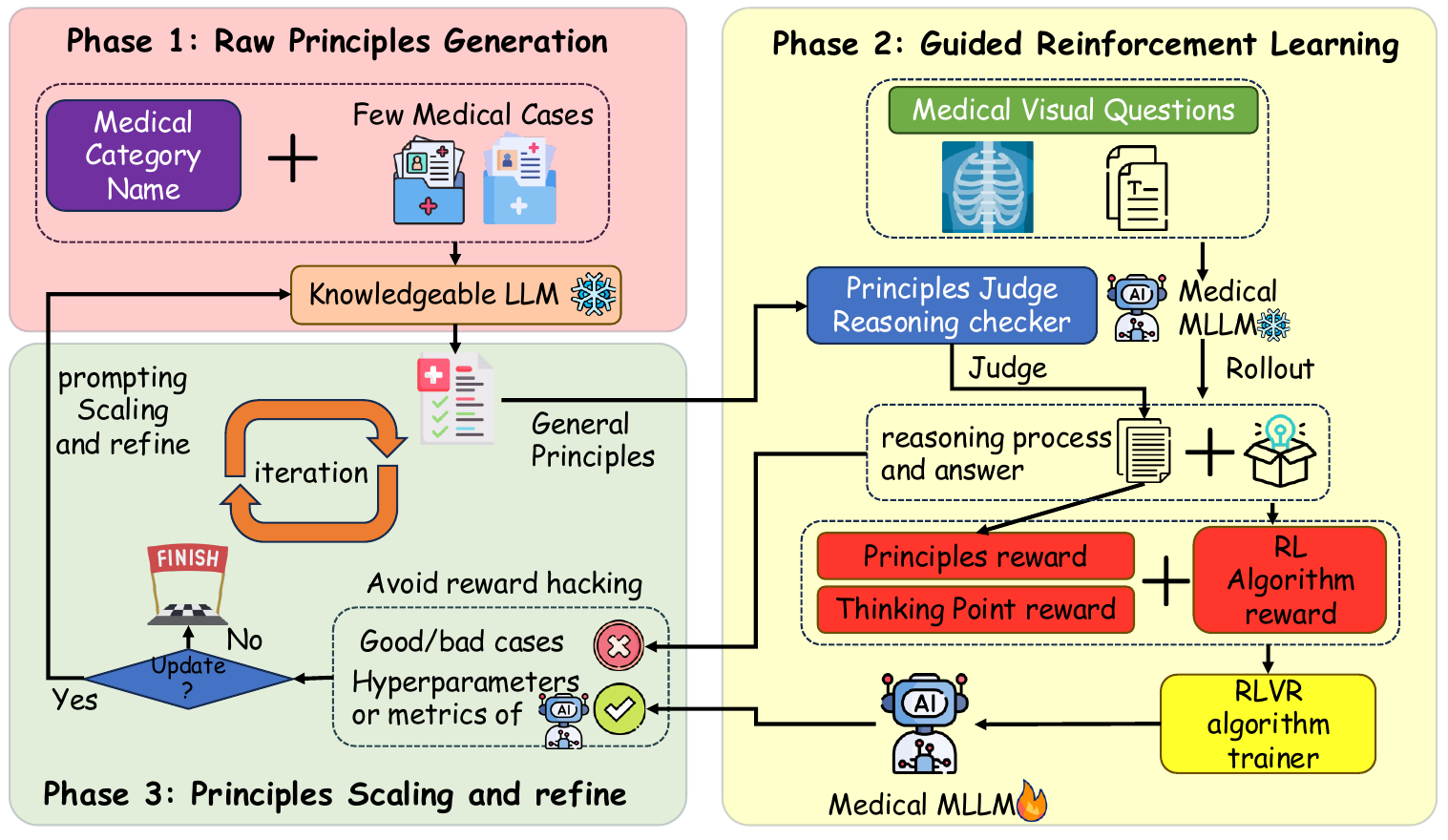} 
    \caption{
    Overview of the \textbf{Evo}lving \textbf{P}rinciple-guided \textbf{I}terative framework (Evo-PI).    
    } 
    \label{framework}
\end{figure*}

\section{Method}
\label{sec:method}
Existing training paradigms lack a principled way to transform abstract domain knowledge into adaptive supervision that evolves with the model.
Consequently, reinforcement learning for multimodal reasoning relies on static scalar rewards that shape outputs but do not guide reasoning.
We propose Evo-PI, an iterative framework that externalizes reasoning principles as explicit supervision signals and continuously refines them via model feedback.

\subsection{Problem Definition}\label{sec:problem}
In medical VQA, an MLLM answers text questions about a medical image. Formally, given a medical MLLM $\mathcal{M}_{\theta}$ with its parameters $\theta$, the input is a set of medical VQA questions. 
Each medical VQA question $q$ consists of a medical image $I$ and a textual question $T$, which can be represented as:
\begin{equation}
\begin{split}
q &= (I, T), \\
I &\in \mathbb{R}^{H \times W \times C}, \\
T &= (w_1, w_2, \dots, w_n), \quad w_i \in \mathcal{V},
\end{split}
\end{equation}
where the textual question $T$ is represented as a sequence of $n$ tokens, with each token $w_i$ drawn from the vocabulary $\mathcal{V}$. $H$, $W$, and $C$ represent the height, width, and channel dimension of the image, respectively.

The model predicts a textual answer $\hat{a}$ to the medical question, represented as a sequence of tokens:
$\hat{a} = (y_1, y_2, \dots, y_m)$, $y_j \in \mathcal{V}$,
where the answer $\hat{a}$ consists of $m$ tokens, and each token $y_j$ is drawn from the same vocabulary $\mathcal{V}$.
A commonly adopted likelihood-based objective for medical VQA is to maximize the conditional probability of the answer given the input, defined as:
\begin{equation}
\begin{split}
&\max_{\theta} \; \log \mathcal{M}_{\theta}(\hat{a} \mid I, T)  \\ 
=& \max_{\theta} \quad \sum_{j=1}^{m} \log \mathcal{M}_{\theta}(y_j \mid y_{<j}, I, T).
\end{split}
\end{equation}
However, this formulation provides no explicit supervision on reasoning quality or clinical validity,
motivating the need for principle-guided supervision in complex medical settings.

\subsection{Overall Framework}
\label{sec:framework_overview}
Evo-PI implements a paradigm for aligning model reasoning with domain principles in a dynamic and iterative manner. 
Rather than treating medical knowledge as a static resource, Evo-PI externalizes knowledge into editable \emph{reasoning principles} that serve as dense supervision signals. 
These principles are continuously refined in response to the model's evolving behaviors, enabling a co-evolutionary learning loop that promotes robust and generalizable reasoning.

As illustrated in Figure~\ref{framework}, Evo-PI consists of three main stages:

\noindent\textbf{(1) Principle Bank Initialization.} 
A frozen, knowledgeable LLM is prompted with few-shot examples and category information to distill a set of initial reasoning principles. 
These principles capture clinically and visually grounded heuristics for interpreting medical images and textual questions, forming the basis for dynamic supervision during training.

\noindent\textbf{(2) Principle-Guided Learning.} 
The backbone MLLM generates answers along with reasoning chains in a rollout process. 
A frozen judge LLM evaluates adherence to the relevant principles, yielding a Principle Reward (for guideline adherence) and a Thinking Point Reward (for logical completeness).
These dense signals are combined with a base RL environment reward to update the model~\cite{GRPO}, shaping not only the final answer but also the underlying logical process.

\noindent\textbf{(3) Principle Evolution.} 
After each iteration, the knowledgeable LLM reviews model behaviors and failure cases to refine and expand the principle set. 
Updated principles are then used in the next iteration, creating a feedback loop where model reasoning and guiding principles co-evolve. 
The process terminates once convergence criteria are satisfied.

By operationalizing supervision at the level of evolving principles rather than static scalar rewards, Evo-PI enables MLLMs to progressively improve reasoning fidelity, generalization, and alignment with expert knowledge.

\subsection{Principle Bank Initialization}
Unlike prior approaches that rely on static heuristics or fixed reward definitions, Evo-PI externalizes domain knowledge into an explicit and editable principle bank.
Inspired by clinical training, where abstract diagnostic principles are distilled and refined through repeated case analysis, this stage converts tacit medical reasoning knowledge into structured language-based supervision. 
Specifically, given a medical VQA dataset with type annotations $\mathcal{C}$, we sample a small anchor set of question-answer pairs $\mathcal{Q}_C$ for each medical type $\text{c}\in \mathcal{C}$. 
Next, Evo-PI employs a frozen Knowledgeable LLM($\mathcal{M}_{K}$) first to generate a set of candidate principles $P_c$ conditioned on a 
medical type name $\text{c}\in \mathcal{C}$ (e.g., MR, CT)
, for all $\text{c}\in \mathcal{C}$. Subsequently, $\mathcal{M}_{K}$ is then prompted with a small random set of question–answer pairs $\mathcal{Q}_C$ drawn for type \text{c} as context, and refines the principles $P$ for coverage and specificity. The initial principle bank is denoted as \(P = \bigcup_{\text{c} \in \mathcal{C}} P_c\). The resulting principles are stored for downstream use. Prompt templates for both initial generation and iterative evolution are provided in Appendix~\ref{principles_prompt}.

We use a text LLM rather than an MLLM at this stage because the goal is medical plausibility, rather than visual grounding. This decouples principle creation from the backbone and modality, requires no shared architecture or pre-training corpus, and avoids alignment constraints.

\subsection{Principle-Guided Reinforcement Learning}
The objective of this stage is to train the medical MLLM to answer visual questions by converting editable principles into reward signals, overcoming fixed heuristic rewards and the lack of knowledge updates in prior work. This formulation allows Evo-PI to move beyond outcome-based supervision, providing dense feedback on how reasoning unfolds, rather than merely whether the final answer is correct.
For each medical VQA instance, the backbone MLLM performs a rollout to produce a reasoning trace and an answer. A frozen judge LLM ($\mathcal{M}_J$) evaluates adherence to the current principle set and returns principle-based rewards. These are merged with the base RL reward, and the composite signal is used to update the parameters of this medical MLLM ($\mathcal{M}_{\theta}$).

Specifically, given a medical VQA input $q=(I,T)$, Evo-PI prompts the learnable medical MLLM $\mathcal{M}_{\theta}$ with an answer format that first elicits a step-wise reasoning trace $rt$. Then a final answer $\hat{a}$ based on this reasoning trace $rt$. The medical MLLM ($\mathcal{M}_{\theta}$) completes one rollout $o=(rt, \hat{a})$ or more $\{o_i\}_{i=1}^{G}$, depending on the chosen RL algorithm. The LLM judge ($\mathcal{M}_J$) scores $o$ against the principles to produce a reward for the principles and a reward for the point of thought, which is combined with the environment reward to update $\theta$.
The detailed prompt templates are available in Appendix \ref{rollout_prompt}.

\paragraph{Principles Judge and Principle Rewards} 
Evo-PI concatenates the principles $P$ from the previous stage into a description paragraph $D=\bigoplus_{i=1}^{n} P_i$, where $\oplus$ denotes the concatenation operation. 
A frozen principles judge LLM ($\mathcal{M}_J$) compares the description paragraph $D$ with the reasoning trace $rt$, and returns a count $c$ of satisfied principles. The principle reward can be formally defined as:
\begin{equation}
    \text{R}_P =
    \begin{cases}
        \dfrac{c}{|P|}, & 0 \leq c \leq |P|, \\
        0, & c < 0 \ \text{or}\ c > |P|,
    \end{cases}
\end{equation}
where the constraint $c>0$ is designed to prevent MLLM from engaging in reward hacking by releasing large amounts of repetitive content in the reasoning traces $rt$. In contrast, $c<0$ reflects the error of the judge from $\mathcal{M}_J$ and is neutralized by clipping. The principle rewards are normalized by $P$ and clip out-of-range values. Simultaneously, the same frozen judge ($\mathcal{M}_J$) verifies, step by step, that the reasoning trace $rt$ supports the final answer $\hat{a}$, analogous to a clinician-style verification workflow. This design transforms abstract principles into verifiable language-based constraints, enabling principled supervision while mitigating reward hacking commonly observed in free-form reasoning rewards.

\paragraph{Thinking Point Reward}
Beyond principle satisfaction, Evo-PI introduces a thinking point reward to encourage structured and verifiable intermediate reasoning.
Rather than evaluating free-form chains of thought, this reward focuses on whether the model explicitly addresses the required reasoning components implied by the principles.

Specifically, Evo-PI extracts the reasoning trace enclosed by \texttt{<think>} tags and parses enumerated reasoning points using regular expressions.
Let $b$ denote the number of distinct reasoning points that are correctly addressed in the trace.
The thinking point reward is defined as:
\begin{equation}
    \text{R}_T =
    \begin{cases}
        \dfrac{b}{|P|}, & 0 \leq b \leq |P|, \\
        0, &  b > |P|.
    \end{cases}
\end{equation}
This reward provides a lightweight and verifiable proxy for reasoning completeness, encouraging the model to articulate clinically meaningful intermediate steps without enforcing a rigid, non-generative template.

\paragraph{RLVR Training Procedure} 
After obtaining reward signals from the Principles Reward (\text{Reward}$_P$) and Thinking Point Reward (\text{Reward}$_T$), Evo-PI treats the backbone MLLM as a policy and optimizes it with Reinforcement Learning with Verifiable Rewards (RLVR) \cite{RLVR}. Specifically, Evo-PI adopts GRPO~\cite{GRPO} and GSPO~\cite{GSPO} as the RLVR trainer. For example, in GRPO, the per-sample scalar return $r_i$ is formed from the verifiable rewards and the environment signal. The normalized per-token advantage is defined as follows:
\begin{equation}
\hat{A}_{i,t} = \frac{r_i - \mu(\{\text{R}_P, \text{R}_T, \text{R}_{RLVR}\})}{\sigma(\text{R}_P, \text{R}_T, \text{R}_{RLVR})},
\end{equation}
where $\text{R}_{RLVR}$ is the original reward defined by the RLVR algorithm (e.g., format reward and accuracy reward in GPRO).
Accordingly, the policy ratio is defined as follows:
\begin{equation}
    r_{i,t}(\theta) =
    \frac{\mathcal{M}_{\theta}\big(o_{i,t} \mid q, o_{i,<t}\big)}
         {\mathcal{M}_{\theta_{\text{old}}}\big(o_{i,t} \mid q, o_{i,<t}\big)}, 
\end{equation}
where $o_i\in $ denotes the $i$-th rollout (i.e., reasoning trace $rt_i$ and answer sequence $\hat{a}_i$) generated by the medical MLLM $\mathcal{M}_{\theta}$ for query $q$, with $G$ total rollouts sampled per query.
The final GRPO objective used to update the $\theta$ combines the principle and thinking point signals with the environment reward to promote stable policy improvement, is given by:
\begin{equation}
\resizebox{0.95\hsize}{!}{$
\begin{aligned}
&\mathcal{J}_{\text{Evo-PI}}(\theta) = \mathbb{E}_{\substack{(q, A) \sim \mathcal{D} \\ \{o_i\}_{i=1}^G \sim \mathcal{M}_{\theta_{\text{old}}}(\cdot \mid q)}} \Bigg[ \quad \frac{1}{G} \sum_{i=1}^{G} \frac{1}{|o_i|} \sum_{t=1}^{|o_i|} \\
&\quad \min\!\Big( r_{i,t}(\theta) \hat{A}_{i,t}, \text{clip}\big(r_{i,t}(\theta), 1 - \varepsilon, 1 + \varepsilon\big) \hat{A}_{i,t} \Big) \Bigg].
\end{aligned}
$}
\end{equation}

\subsection{Principle Evolution}
As the backbone model improves, static supervision inevitably becomes misaligned with its evolving reasoning capabilities.
Principle evolution enables supervision to scale in abstraction and coverage, maintaining alignment throughout training.
Phase three consists of three core components to adapt the set of principles by controlling the iteration, scaling the principles, and refining them, rather than relying on fixed rewards or static knowledge during training.

\noindent \textbf{Iteration and Generalization Control }
To prevent overfitting to cases produced during training and limit drift from medical knowledge, we cap the number of loop passes and apply early stopping on a held-out validation set.

\noindent \textbf{Balance of Exploitation and Exploration }
We track token-level policy entropy during the process of reinforcement learning. Training terminates typically under two conditions: \textit{Entropy Collapse}~\cite{entropy_collapse} and \textit{Abnormal Entropy Increase}. 
Entropy collapse is a common phenomenon indicating a reduction in the MLLM's exploration space, as the model tends to become more certain of its own inferences, also known as exploitation over exploration.
High entropy indicates that the policy prioritizes exploration to discover solutions, whereas low entropy signifies a focus on consistently exploiting optimal actions.
In Evo-PI, medical MLLMs serve as policy models that are updated by the RLVR trainer, where the balance of exploring diverse case patterns with reliably solving individual cases makes entropy a suitable stopping signal.
Because principles are embedded in the judge and never exposed to the MLLM, reward hacking is curtailed, though mild entropy rises can occur.
In practice, Evo-PI terminates training if entropy falls to a collapse threshold or exceeds the baseline recorded after the first MLLM update.

\noindent \textbf{Scaling and Refinement of Principles } If termination criteria are not met, a frozen Knowledgeable LLM receives the current principles from the latest round and prompts it to scale and refine these principles. 
The update increases coverage while keeping rules abstract, concise, and compositional by merging redundant rules, pruning low-utility rules, and adding new numbered rules only for recurring failure modes.
Prompts for this step are given in Appendix~\ref{principles_update}.
\section{Experiments}
\label{sec:experiments}

We conduct quantitative and qualitative experiments to evaluate Evo-PI.

\subsection{Experimental Setting}

\paragraph{Datasets}
We conduct experiments on the OmniMedVQA dataset~\cite{OmniMedVQA}, which is used in Med-R1 and designed for the medical VQA task. OmniMedVQA spans eight imaging modalities: Computed tomography (CT), dermoscopy (DER), fundus photography (FP), microscopy images (MI), magnetic resonance imaging (MR), optical coherence tomography (OCT), ultrasound (US), and X-ray. After removing duplicate cases, we split the data into training, testing, and validation sets in an 8:1:1 ratio.
Table~\ref{statistics} summarizes the statistics of these eight benchmark datasets.

\begin{table}[htp]
\centering
   
\setlength\tabcolsep{11pt}
\resizebox{\linewidth}{!}{
    \begin{tabular}{lrrrrrrrr}
    \toprule
    Dataset& CT & DER & FP & MI & MR & OCT & US & X-ray \\ \midrule
    Training & 12,647 & 5,343 & 4,318 & 4,544 & 25,501 & 3,716 & 8,792 & 6,332 \\
    Validation & 1,581 & 668 & 540 & 568 & 3,188 & 465 & 1,099 & 792 \\
    Test & 1,581 & 668 & 540 & 568 & 3,188 & 465 & 1,100 & 792 \\
    Total & 15,809 & 6,679 & 5,398 & 5,680 & 31,877 & 4,646 & 10,991 & 7,916 \\
    
    \bottomrule
    \end{tabular}
    }
    \caption{Statistics of the datasets.}
    \label{statistics}
\end{table}

\paragraph{Hyperparameter Settings}
Evo-PI employs a consistent set of hyperparameters across all eight datasets. The default iteration number is three, with one training epoch in each iteration. 
The batch size is 256 for all backbone medical MLLMs.
For group-based RL algorithms (such as GRPO), the rollout number defaults to 8. 
For RL algorithms using cilp-higher (e.g., GSPO), Evo-PI set the \texttt{clip\_ratio\_low}=0.0003 and \texttt{clip\_ratio\_high}=0.0004.

We recommend knowledgeable and judge LLMs with at least 7B parameters. In our setup, GPT-4o-mini\footnote{\url{https://platform.openai.com/docs/models/gpt-4o-mini}} serves as the knowledgeable LLM for principle generation, scaling, and refinement, and Qwen2.5-VL-7B\footnote{\url{https://huggingface.co/Qwen/Qwen2.5-VL-7B-Instruct}} serves as the judge LLM to evaluate the responses from medical MLLM.
We implemented parallelism using FSDP on the Verl training framework. All backbone models are trained on  4$\times$H100 80GB SXM GPUs, and the judge LLM runs on separate 2$\times$H100 80GB GPU. On the largest MR dataset, the total runtime is within 20 hours.

\paragraph{Baselines and Evaluation Metrics}
We evaluate Evo-PI against two strong medical VQA backbones, HuatuoGPT-Vision ~\cite{HuatuoGPT-Vision} and Med-R1~\cite{Med-R1}, both state-of-the-art on medical VQA. For each backbone, we report Accuracy before and after applying Evo-PI under identical decoding and data splits. 
We also conduct a horizontal comparison across base MLLM variants of varying sizes corresponding to the backbones’ underlying models.

\begin{table*}[bth]
    \footnotesize
	\centering
    \tabcolsep=1pt
        
    \resizebox{0.95\linewidth}{!}{
	\begin{tabular}{lccccccccc}
		\toprule
		Dataset& CT & DER & FP & MI & MR & OCT & US & X-ray& Average \\
        
        \midrule

        \textbf{Qwen 2-VL-2B [1]} & 0.4023 & 0.4177 & 0.4685 & 0.4208 & 0.4410 & 0.3828 & 0.3864 & 0.4823 & 0.4252\\
        \textbf{Qwen 2-VL-7B [2]} &0.6818 & 0.5719 & 0.7444 & 0.6250 & 0.5452 & 0.6323 & 0.3827 & 0.7197 & 0.6129\\
        \textbf{Qwen 2-VL-72B [3]} & 0.6797 & 0.6531 & 0.7258 & 0.6784  & 0.6939 & 0.7276 & 0.5139 & 0.7211 & 0.6805\\
        \textbf{Qwen 2.5-VL-3B [4]}  & 0.7103 & 0.6078 & 0.6981 & 0.5986 & 0.5364 & 0.6839 & 0.4409 & 0.7462 & 0.6278\\
        \textbf{Qwen 2.5-VL-7B [5]}  & 0.6736 & 0.7006 & 0.7056 & 0.6004 & 0.5533 & 0.5570 & 0.3355 & 0.7563 & 0.6103\\
        \textbf{Qwen 2.5-VL-72B [6]}  & 0.6618 & 0.6975 & 0.7104 & 0.6937 & 0.6364 & 0.6922 & 0.6985 & 0.7981 & 0.6771\\

        \midrule
        \textbf{LLaVA-Med [7]}      & 0.1869 & 0.4495 & 0.3903 & 0.3329 & 0.2747 & 0.3461 & 0.2988 & 0.3068 & 0.3233 \\
        \textbf{RadFM [8]}          & 0.2756 & 0.3921 & 0.3686 & 0.2797 & 0.2406 & 0.3280 & 0.1657  & 0.3095  & 0.2950 \\
        \textbf{Med-Flamingo [9]}   & 0.3128 & 0.4856 & 0.4126 & 0.3003 & 0.2634 & 0.2516 & 0.3169 & 0.4401 & 0.3429 \\
        \textbf{MedVInT [10]}        & 0.4074 & 0.2911 & 0.3184 & 0.3202 & 0.4310 & 0.2326 & 0.4126 & 0.5510 & 0.3705 \\
        
        \midrule        
        \textbf{HuatuoGPT-Vision [11] (Base [5])} & 0.6534 & 0.6841 & 0.7630 & 0.7130 & 0.6866 & 0.7763 & 0.4818 & 0.8005 & 0.6948\\
        \textbf{Evo-PI (Base [11] + GSPO)} & 0.8797 & 0.9021 &  \textbf{0.9369} &  \textbf{0.9580} &  0.9138 & 0.9824 & 0.9109 & \textbf{0.9367} & 0.9295 \\
        \textbf{Evo-PI (Base [11] + GRPO)} & 0.8797 & 0.9021 &  \textbf{0.9369} &  \textbf{0.9580} & 0.9418 & 0.9824 & 0.9109 & \textbf{0.9367} &  0.9295 \\
        \textbf{Relative Gains (\%)}  &    {22.63}\%\textcolor{blue}{$\uparrow$}  &   {23.70}\%\textcolor{blue}{$\uparrow$}  &   {18.91}\%\textcolor{blue}{$\uparrow$}  &   {24.50}\%\textcolor{blue}{$\uparrow$}  &   {25.52}\%\textcolor{blue}{$\uparrow$}  &   {20.61}\%\textcolor{blue}{$\uparrow$} &   {47.28}\%\textcolor{blue}{$\uparrow$} &   {13.62}\%\textcolor{blue}{$\uparrow$} &   {24.59}\%\textcolor{blue}{$\uparrow$}\\
        
        \midrule
        
        \textbf{Med-R1 [12] (Base [1])} & 0.7160 & 0.8338 & 0.9019 & 0.7447 & 0.5144 & 0.8946 & 0.7773 & 0.7854 &0.7710\\
        \textbf{Evo-PI (Base [12] + GSPO)} & \textbf{0.9628} &  \textbf{0.9334} & 0.9081 & 0.8739 & 0.9298 & \textbf{0.9934} & \textbf{0.9982} & 0.9038 & 0.9391\\
        \textbf{Evo-PI (Base [12] + GRPO)} & 0.9676 & 0.9319 &  0.9195 & 0.8720 & \textbf{0.9462} & 0.9890 & \textbf{0.9982} & 0.9101 & \textbf{0.9412}\\
        \textbf{Relative Gains (\%)}  &    {25.16}$\%$\textcolor{blue}{$\uparrow$}  &   {9.96}$\%$\textcolor{blue}{$\uparrow$}  &   {1.76}$\%$\textcolor{blue}{$\uparrow$}  &   {12.92}$\%$\textcolor{blue}{$\uparrow$}  &   {43.18}$\%$\textcolor{blue}{$\uparrow$}  &   {9.88}$\%$\textcolor{blue}{$\uparrow$} &   {22.09}$\%$\textcolor{blue}{$\uparrow$} &   {12.47}$\%$\textcolor{blue}{$\uparrow$} &   {17.18}$\%$\textcolor{blue}{$\uparrow$} \\

        \midrule
        \textbf{Relative Gains (Task)} &    {23.89}$\%$\textcolor{blue}{$\uparrow$}  &   {16.83}$\%$\textcolor{blue}{$\uparrow$}  &   {10.34}$\%$\textcolor{blue}{$\uparrow$}  &   {18.71}$\%$\textcolor{blue}{$\uparrow$}  &   {34.35}$\%$\textcolor{blue}{$\uparrow$}  &   {15.24}$\%$\textcolor{blue}{$\uparrow$} &   {34.68}$\%$\textcolor{blue}{$\uparrow$} &   {13.05}$\%$\textcolor{blue}{$\uparrow$} &   {20.88}$\%$\textcolor{blue}{$\uparrow$} \\

		\bottomrule
    \end{tabular}
    }
    \caption{Overall comparison on the medical visual question answering task.
    \textbf{Bold} indicates the best results.
    HuatuoGPT-Vision is based on Qwen 2.5-VL-7B. Med-R1 comprises eight modality-specific submodels for the eight medical modalities, each based on Qwen 2-VL-2B.
    All our experimental results are averaged over three independent runs.
    }
	\label{results_for_all}
\end{table*}

\subsection{Overall Performance Comparison}
The main quantitative results for the medical VQA task are reported in Table~\ref{results_for_all}.
Several consistent trends can be observed.
First, Evo-PI consistently improves all backbone MLLMs
across the eight imaging modalities and different RL training algorithms.
Specifically, Evo-PI achieves accuracy improvements ranging from 10.34\% to 34.68\%,
demonstrating robust effectiveness under diverse medical imaging conditions.
Second, Evo-PI remains effective when applied to strong medical VQA backbones.
On HuatuoGPT-Vision (7B), Evo-PI improves the backbone performance by an average of 24.59\%.
Moreover, on Med-R1 models trained with the same GRPO optimizer,
Evo-PI yields an average improvement of 17.18\%,
indicating that the proposed framework complements existing RLVR training schemes.
Finally, on OCT and ultrasound datasets,
Evo-PI–enhanced models surpass 99.3\% accuracy for the first time,
suggesting that principle-guided supervision can push medical VQA performance toward clinically reliable regimes.

\subsection{Impact of Iteration Design}
In Evo-PI, the guiding principles are updated at each iteration. We investigate the respective effects of each factor under an iterative mechanism.

\begin{figure}[htb]
    \centering
    \includegraphics[width=\linewidth]{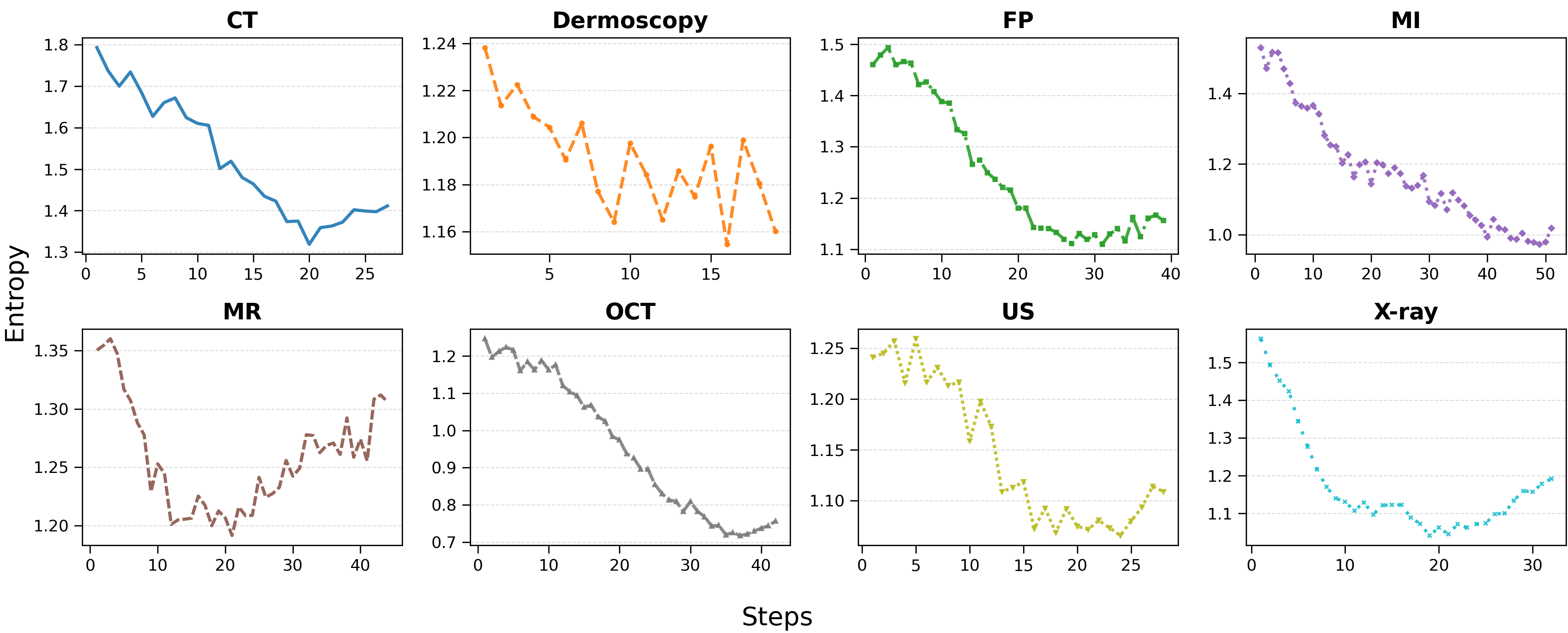}
    \caption{
    Entropy dynamics for eight independent training runs, each with its own step axis.
    Each subplot corresponds to a modality: CT, DER, FP, MI, MR, OCT, US, and X-ray. 
    All backbones are from the Med-R1 series, and all runs use GRPO.
    }
    \label{fig:entropy_subplots}
\end{figure}

\paragraph{Principles Cases in Iteration}
At each iteration, the knowledgeable LLM initializes or updates the current set of principles, enabling direct observation of how they evolve.

We observe two general phenomena.
(i) When the initial set is incomplete, the knowledgeable LLM refines or decomposes existing principles while introducing additional ones that better meet the task requirements.
(ii) When the initial set is already reasonably comprehensive, the knowledgeable LLM performs fine-grained edits, improving individual principles one by one.

Taking the principles for X-ray as an example, we demonstrate how they are generated and how they are iteratively scaled and refined in Appendices~\ref{principles_prompt} and ~\ref{principles_update}, respectively.

\paragraph{Entropy Analysis in Iteration}
Entropy quantifies the exploration–exploitation balance of a policy, offering an interpretable lens on learning dynamics \cite{entropy,entropy_collapse}. High entropy typically signals exploratory behavior to discover new solutions, whereas low entropy indicates confident, exploitative behavior with more consistent action selection.

In Evo-PI, medical MLLMs are optimized as policies under an RLVR trainer. They must balance exploring diverse case patterns with reliably solving individual cases. This motivates using entropy-based stopping: we terminate an iteration when entropy evidences either collapse (i.e., over exploitation) or abnormal increase (i.e., unstable exploration), marking ineffective learning progress. 
As shown in Figure~\ref{fig:entropy_subplots}, entropy decreases across all modalities, with low amplitude fluctuations near the end of training that reflect the evolving exploration–exploitation balance under Evo-PI.

\paragraph{Ablation on Principles Gains and Iteration Gains}
The gains derived from principles and iteration demonstrate how the principles within iterative process exert their influence in Evo-PI.
We conducted ablation experiments on the results for Med-R1 when using GRPO as the RLVR algorithm.

We find that applying the principles yields an average improvement of 13.48\%. On top of this, introducing iteration provides an additional performance boost of around 3.7\%.
Detailed ablations are provided in Appendix~\ref{ablation}.

\subsection{Qualitative Studies}
We conducted tests using Evo-PI across all modalities and observed marked improvements in reasoning quality: post-training traces become more coherent, clinically grounded, and logically ordered. We attribute the performance gains primarily to this enhancement in stepwise reasoning rather than superficial answer heuristics. A representative CT case is shown below:

\newpage
\begin{tcolorbox}[
    colback=gray!5, 
    colframe=gray!60!black,
    title=Case Study: CT Scan,
    fonttitle=\small\sffamily,
    boxrule=0.8pt,
    sharp corners,
    left=1mm, right=1mm, top=1mm, bottom=1mm,
    breakable
]
\centering
\includegraphics[width=0.2\linewidth]{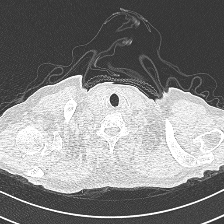} \\[6pt]

\raggedright 

\textbf{Problem:} Is there any perceptible anomaly present in this picture? \\
\quad A)No \quad B)Yes

\textbf{Ground Truth:} A \\
\quad \\
\textbf{Med-R1's answer}:

\texttt{\textless think\textgreater}
The image description
A grayscale image of a cross-sectional view of a human head, showing the brain, skull, and surrounding tissues. The image appears to be a medical scan, possibly an MRI or CT scan, with various shades of gray representing different tissue densities.
\texttt{\textless/think\textgreater}
\texttt{\textless answer\textgreater B\textless/answer\textgreater} 
\quad \\

\textbf{Evo-PI's answer (evolved)}:

\texttt{\textless think\textgreater}
1. The image provided is a CT scan of the neck, which is typically used to visualize soft tissues and structures within the neck.

2. The scan shows a cross-sectional view of the neck, including the thyroid gland, which is a critical structure in the neck.

3. The thyroid gland appears to be in its normal position, with no visible signs of enlargement or abnormal shape.

4. There are no visible masses, calcifications, or other abnormalities that would suggest a pathological condition.

5. The surrounding structures, such as the trachea and carotid arteries, appear to be in their normal anatomical positions.

6. The image does not show any signs of inflammation, infection, or other conditions that would typically be indicated by abnormal findings on a CT scan.

Based on the above observations, there does not appear to be any perceptible anomaly present in the image.

\texttt{\textless/think\textgreater}
\texttt{\textless answer\textgreater A\textless/answer\textgreater}

\end{tcolorbox}

From this example, the Evo-PI enhanced answer exhibits stronger logical coherence and makes fuller use of CT-specific knowledge than the original Med R1 output, leading to the correct solution. Additional case studies are provided in Appendix~\ref{Additional_case_studies}.

\section{Related Work}
\label{sec:related_work}

\paragraph{General MLLMs and Medical MLLMs}
Recent years have seen the emergence of large-scale MLLMs, such as GPT-4o~\cite{GPT4} and the Gemini series~\cite{gemini2.5}. These models acquire strong cross-modal reasoning by leveraging vast data collections,
but often struggle to exhibit reliable, clinically grounded reasoning in specialized medical tasks.
Medical MLLMs address this gap, typically through SFT on expert-annotated datasets~\cite{HuatuoGPT-Vision, lu2025harnessing}. While effective, such approaches incur substantial annotation costs, creating a major bottleneck.  
To mitigate data scarcity, recent works have explored RL for post-training alignment~\cite{MedVLM-R1, Med-R1, PathVLM-R1}. However, these methods typically treat medical knowledge as static supervision signals (e.g., datasets or fixed reward heuristics),
rather than as explicit, evolving principles that actively guide the learning process.
As a result, they largely overlook structured medical knowledge during alignment,
despite the demonstrated effectiveness of knowledge-aware judging in general LLM settings~\cite{LLM_as_a_Judge}.
This gap highlights the need to rethink how medical expertise is integrated into the alignment process.

\paragraph{Reinforcement Learning Algorithms for Post-Training}
State-of-the-art alignment commonly employs RL algorithms such as PPO~\cite{PPO} and GRPO~\cite{GRPO}, with subsequent variants like GSPO~\cite{GSPO} addressing challenges including training instability and entropy collapse~\cite{entropy_collapse}. A common limitation across these techniques is their reliance on reward functions with fixed, pre-defined strategies. Such rigidity leaves models susceptible to reward hacking~\cite{Reward_Hacking}, particularly in knowledge-intensive domains like medicine, where they may exploit surface-level heuristics of the reward scheme rather than acquiring deeper, clinically aligned reasoning skills.  
These limitations motivate our central hypothesis:
robust medical reasoning requires reward mechanisms that are \emph{dynamic, knowledge-driven, and grounded in explicitly articulated principles}
that co-evolve with the model during training.

\section{Conclusion}
\label{sec:conclusion}
We present Evo-PI, a principle-guided framework for medical VQA that treats reasoning principles as evolvable supervision signals.
By combining principle-guided reinforcement learning with iterative refinement, Evo-PI aligns model reasoning with clinical knowledge.
Experiments demonstrate consistent performance gains and robustness across modalities and backbones.
Evolving supervision offers a promising direction for reliable, interpretable, and expert-aligned reasoning in MLLMs.

\section{Acknowledgments}
This research was supported in part by the Young Scientists Fund of the National Natural Science Foundation of China (Grant No.~32500997, S.~Li), and in part by Beijing Renyixun Health Technology Co., Ltd.
This research is also partly supported by the Shenzhen Science and Technology Program No. SYSPG20241211173609009, and National Science and Technology Council (NSTC), Taiwan (Grants: NSTC-114-2222-E-A49-004, and NSTC-114-2639-E-A49-001-ASP).

\section*{Limitations}

Despite its effectiveness, the proposed Evo-PI framework is subject to several limitations that suggest important directions for future research.

First, Evo-PI relies on the iterative generation and refinement of guiding principles, whose quality is inherently influenced by the reasoning capacity and domain knowledge of the underlying judge model. Although we observe stable improvements across different backbones and reinforcement learning algorithms, suboptimal or overly generic principles may still limit performance gains in highly specialized or rare medical scenarios. Improving the controllability and specialization of principle induction remains an open challenge.

Second, our framework introduces additional computational overhead due to the principle evolution and reward construction process. While Evo-PI does not require modifying backbone model architectures, the iterative interaction among generation, judging, and reinforcement learning increases training cost compared to standard supervised fine-tuning. Exploring more efficient principle updating strategies or lightweight judge models would be beneficial for large-scale or resource-constrained deployments.

Third, our empirical evaluation focuses on medical VQA tasks with well-defined visual and textual supervision. Although this setting provides a rigorous testbed for principled reasoning under high-stakes conditions, the generalizability of Evo-PI to other multimodal reasoning tasks (e.g., longitudinal clinical decision-making or non-medical domains) has not yet been systematically studied. Extending the framework to broader reasoning scenarios and assessing its domain-agnostic properties constitute promising directions for future work.

Importantly, Evo-PI does not require perfectly accurate judges; principles are evaluated in a relative and iterative manner, which empirically provides robustness to moderate judging noise.

\bibliography{ref}
\newpage
\appendix
\section{Appendix}
\label{sec:appendix}

\subsection{The Guideline of Appendix}
The appendix is organized as follows:

\begin{itemize}
    \item In section ~\ref{Ethics}, we introduce Ethics Statement of our paper.
    \item In section ~\ref{Reproducibility_statement}, we introduce Reproducibility Statement of our paper.
    \item In section ~\ref{Risks}, we introduce Potential Risks of our paper.
    \item In section ~\ref{LLM_usage}, we introduce how we use the LLM in our work as required.
    \item In section ~\ref{principles_prompt}, we introduce how we generate and scaling principles for each sub-modality of medical questions.
    \item In section ~\ref{principles_update}, we introduce how these principles are scaling and refined.
    \item In section ~\ref{rollout_prompt}, we introduce the post-training prompt to control the rollout process.
    \item In section ~\ref{ablation}, we further introduce some ablation study about the impact of the iterations and the principles.
    \item In section ~\ref{Additional_case_studies}, we introduce the case study across different datasets.
\end{itemize}

\subsection{Ethics Statement}\label{Ethics}
All datasets used in this research are publicly available and were sourced from previous studies that have undergone appropriate ethical review. Our work did not involve the collection of any new data from human subjects. We have adhered to all data usage agreements and licenses associated with these pre-existing datasets.

\subsection{Reproducibility Statement}\label{Reproducibility_statement}
We are committed to making our research reproducible. All datasets used in this study are publicly available, and we provide detailed descriptions and sources in our experimental setup section.

\subsection{Potential Risks}\label{Risks}
We do not identify significant risks associated with our approach, as it relies solely on open-source data and models and follows standard research practices.

\subsection{Usage of LLM}\label{LLM_usage}

This paper primarily employs large language models (LLMs) to refine the overall quality of writing, with a particular focus on eliminating incorrect expressions, minimizing grammatical errors, and enhancing clarity, coherence, and readability to ensure the text meets standards of ACL.

\subsection{Prompts for Generating and Scaling Raw Principles}\label{principles_prompt}

In this section, we demonstrate how we generate and update principles. Using the principles corresponding to X-ray cases as an example, Evo-PI utilizes five cases from the training set and then instructs the knowledgeable LLM to complete the generation of the principles.

When principles are updated, Evo-PI will provide the previously used principles and streamline and scale them.

\begin{lstlisting}[
    caption={Raw Principles Generation Prompt},
    basicstyle=\footnotesize\ttfamily,
    breaklines=true,
    columns=fullflexible,
    breakatwhitespace=false,
    keepspaces=true,
    frame=single,
    showstringspaces=false
]
list some principles for these key medical reasoning tasks "X-Ray"

Here are the sample questions and answers:
    {
        "image": "Images/RadImageNet/bladder_pathology/abd132420.png",
        "problem": "What is the abnormality present in the image? A)Spinal cord injury B)Ovarian cyst C)Bladder pathology D)Liver cirrhosis",
        "solution": "<answer> C </answer>"
    },
    {
        "image": "Images/RadImageNet/normal/abd-normal028416.png",
        "problem": "Is there any deviation or anomaly observed in this image? A)No B)Yes.",
        "solution": "<answer> A </answer>"
    },
    {
        "image": "Images/RadImageNet/post_op/abd000807.png",
        "problem": "What type of abnormality is present in this image? A)Foreign body reaction B)Post-operative changes C)Infection D)Fracture site healing",
        "solution": "<answer> B </answer>"
    },
    {
        "image": "Images/RadImageNet/normal/abd-normal056990.png",
        "problem": "Is anything out of the ordinary evident in this image? A)No B)Yes.",
        "solution": "<answer> A </answer>"
    },
    {
        "image": "Images/RadImageNet/interstitial_lung_disease/lung044245.png",
        "problem": "What type of abnormality is present in this image? A)Interstitial lung disease B)Pulmonary hypertension C)Asthma D)Pleural effusion",
        "solution": "<answer> A </answer>"
    },
and save it as a string list in Python.
\end{lstlisting}

\begin{lstlisting}[
    caption={Principles Updating Prompt},
    basicstyle=\footnotesize\ttfamily,
    breaklines=true,
    columns=fullflexible,
    breakatwhitespace=false,
    keepspaces=true,
    frame=single,
    showstringspaces=false
]
principles_prompt = [
    {"role": "system", "content": "You are helpful AI system in medical.\n\n"},
    {"role": "user", "content": f"Here are some key general principles for medical reasoning tasks involving {question_type}:\n\n"},
    {"role": "user", "content": f"{principles}\n\n"},
    {"role": "user", "content": "The principles should keep general and abstract, and as short as possible. If necessary, extend and merge the list of principles with new numbered principles; otherwise keep current principles unchanged. The listed principles should not exceed ten points The merged principles list:"}
]
\end{lstlisting}

\subsection{Prompt for Generating Raw Principles}\label{principles_update}

In this section, we will use the principles of X-ray as an example to demonstrate how these principles are specifically updated.

\begin{tcolorbox}[title=Principles for X-ray in iteration 1, breakable]
1. Recognize common imaging signatures. X-rays typically show high-contrast grayscale images. Bone appears white, air (such as in the lungs) appears dark, and soft tissue is various shades of gray. Look for clear bony landmarks like ribs, spine, and clavicles.\\

2.Identify typical anatomical projections. Chest X-rays are often captured in posteroanterior (PA) or anteroposterior (AP) views. These images generally show the patient in an upright posture with visible lungs, diaphragm, and heart shadow.\\

3.Use process of elimination. Eliminate options that do not involve imaging (e.g., blood test, EKG). MRI images typically show high soft tissue contrast but have less visible bone detail. Ultrasound images have a grainy texture and do not show bones clearly. PET scans often include colorful overlays or are fused with CT, and do not resemble plain grayscale images.\\

4.Integrate clinical context. X-rays are commonly used as a first-line imaging tool for evaluating chest-related symptoms or conditions such as pneumonia, pneumothorax, or congestive heart failure.\\

5.Pay attention to image characteristics. X-rays are uniform grayscale images without the detailed cross-sectional anatomy seen in CT or MRI. They do not include high-resolution soft tissue differentiation or 3D rendering.\\

6.Rely on pattern recognition rather than memorization. Become familiar with the visual features that distinguish different imaging modalities. Recognize typical X-ray features through exposure to many examples.
\end{tcolorbox}

\begin{tcolorbox}[title=Principles for X-ray in iteration 2, breakable]
1. **Recognize Imaging Signatures:** Understand the high-contrast nature of X-rays, where bones appear white, air is dark, and soft tissues are gray.\\

2. **Understand Anatomical Projections:** Be familiar with common X-ray views like posteroanterior (PA) and anteroposterior (AP) to properly interpret anatomical structures.\\

3. **Differentiate Imaging Modalities:** Identify X-rays by their grayscale images and lack of high-resolution soft tissue contrast.\\

4. **Integrate Clinical Context:** Use X-rays in the context of clinical presentation to assess conditions like pneumonia, pneumothorax, and heart failure.\\

5. **Acknowledge Limitations:** Be aware that X-rays do not provide detailed cross-sectional or 3D images.\\

6. **Employ Pattern Recognition:** Develop proficiency through practice and focus on identifying typical X-ray patterns.\\

7. **Enhance Diagnostic Accuracy:** Combine X-ray findings with clinical information for a comprehensive evaluation.
\end{tcolorbox}

\begin{tcolorbox}[title=Principles for X-ray in iteration 3, breakable]
1. **Recognize Imaging Patterns:** Identify distinctive grayscale patterns for bones, air spaces, and soft tissues in X-rays.\\

2. **Understand Projections:** Be aware of common anatomical projections like PA and AP views for accurate image assessment.\\

3. **Differentiate Modalities:** Distinguish features of X-rays from those of MRI, CT, and ultrasound.\\

4. **Contextualize Clinically:** Integrate clinical data to enhance X-ray interpretation, aiding in diagnosis of conditions like pneumonia.\\

5. **Acknowledge Limitations:** Recognize the limitations of X-rays, such as limited soft tissue detail and no 3D rendering.\\

6. **Apply Pattern Recognition:** Develop the skill to efficiently recognize and interpret common X-ray findings.\\

7. **Ensure Systematic Approach:** Adopt a systematic approach to evaluating X-rays to ensure comprehensive analysis.\\

8. **Correlate with Findings:** Relate X-ray observations with clinical examination and laboratory results for a holistic view.\\

9. **Prioritize Safety:** Maintain awareness of radiation exposure and ensure justified use of X-ray imaging.\\

10. **Stay Informed:** Keep updated with evolving X-ray technology and guidelines for best practices.
\end{tcolorbox}

\subsection{Prompt for Generating Responses from Medical MLLMs}\label{rollout_prompt}

For this section, we directly adopted the prompt provided on med-r1, thereby eliminating any sensitivity associated with the prompt.

\begin{tcolorbox}[title=rollout prompt, breakable]
QUESTION\_TEMPLATE = "{Question} First output the thinking process in \texttt{\textless think\textgreater B\textless/think\textgreater}  and final choice (A, B, C, D ...) in \texttt{\textless answer\textgreater \textless/answer\textgreater}  tags."

\end{tcolorbox}

\begin{table*}[bth]
    \small
	\centering
    \tabcolsep=1pt
    \resizebox{\linewidth}{!}{
	\begin{tabular}{lccccccccc}
		\toprule
		Dataset& CT & DER & FP & MI & MR & OCT & US & X-ray& Average \\

        \midrule
        
        \textbf{Med-R1 [12] (Base [1])} & 0.7160 & 0.8338 & 0.9019 & 0.7447 & 0.5144 & 0.8946 & 0.7773 & 0.7854 &0.7710\\
        \textbf{Evo-PI (Fix principles)} &0.9278	&0.9243	&0.9140	&0.8343	&0.8174	&0.9505	&0.9955	&0.8747	&0.9048\\
        \textbf{Evo-PI (Iteration principles)} & 0.9676 & 0.9319 &  0.9195 & 0.8720 & {0.9462} & 0.9890 & {0.9982} & 0.9101 & {0.9412}\\

        \midrule

        \textbf{Relative Gains from principles only (\%)}  &    {21.18}$\%$\textcolor{blue}{$\uparrow$}  &   {9.21}$\%$\textcolor{blue}{$\uparrow$}  &   {1.21}$\%$\textcolor{blue}{$\uparrow$}  &   {9.14}$\%$\textcolor{blue}{$\uparrow$}  &   {30.30}$\%$\textcolor{blue}{$\uparrow$}  &   {6.03}$\%$\textcolor{blue}{$\uparrow$} &   {21.82}$\%$\textcolor{blue}{$\uparrow$} &   {8.93}$\%$\textcolor{blue}{$\uparrow$} &   {13.48}$\%$\textcolor{blue}{$\uparrow$} \\
        \textbf{Relative Gains from iteration only (\%)}  &    {3.98}$\%$\textcolor{blue}{$\uparrow$}  &   {0.76}$\%$\textcolor{blue}{$\uparrow$}  &   {0.55}$\%$\textcolor{blue}{$\uparrow$}  &   {3.77}$\%$\textcolor{blue}{$\uparrow$}  &   {12.88}$\%$\textcolor{blue}{$\uparrow$}  &   {3.85}$\%$\textcolor{blue}{$\uparrow$} &   {0.27}$\%$\textcolor{blue}{$\uparrow$} &   {3.54}$\%$\textcolor{blue}{$\uparrow$} &   {3.70}$\%$\textcolor{blue}{$\uparrow$} \\

        \textbf{Relative Gains total (\%)}  &    {25.16}$\%$\textcolor{blue}{$\uparrow$}  &   {9.96}$\%$\textcolor{blue}{$\uparrow$}  &   {1.76}$\%$\textcolor{blue}{$\uparrow$}  &   {12.92}$\%$\textcolor{blue}{$\uparrow$}  &   {43.18}$\%$\textcolor{blue}{$\uparrow$}  &   {9.88}$\%$\textcolor{blue}{$\uparrow$} &   {22.09}$\%$\textcolor{blue}{$\uparrow$} &   {12.47}$\%$\textcolor{blue}{$\uparrow$} &   {17.18}$\%$\textcolor{blue}{$\uparrow$} \\

		\bottomrule
    \end{tabular}
    }
     \caption{Overall ablation study 
    }

	\label{results_for_ablation}
\end{table*}

\subsection{Detailed Ablation Study}\label{ablation}

In this section, we present the results of ablation experiments using Med-R1 as the backbone MLLM. 
We observe that the average gain achieved by applying the principles reaches 13.48\%. Building upon this foundation, incorporating iteration can further enhance performance by approximately 3.7\%.
On datasets like DER and US, which are relatively principle-dependent, the gains from starting with good initial principles are the greatest.

\subsection{Additional Qualitative Studies}\label{Additional_case_studies}

For this section, we present case studies across various sub-datasets.
\begin{tcolorbox}[title=Sample of CT case, breakable]
\centering
\includegraphics[width=0.2\linewidth]{figs/ct_case.png} \\[6pt]

\raggedright 

\textbf{Problem:} Is there any perceptible anomaly present in this picture? \\
\quad A)No \quad B)Yes

\textbf{Ground Truth:} A \\
\quad \\

\raggedright 
\textbf{Med-r1's answer}:

\texttt{\textless think\textgreater}
The image description
A grayscale image of a cross-sectional view of a human head, showing the brain, skull, and surrounding tissues. The image appears to be a medical scan, possibly an MRI or CT scan, with various shades of gray representing different tissue densities.
\texttt{\textless/think\textgreater}
\texttt{\textless answer\textgreater B\textless/answer\textgreater} 

\quad \\
\textbf{Evo-PI's answer (during iteration)}:

\texttt{\textless think\textgreater}
The image provided is a CT scan of the neck, which is typically used to visualize the soft tissues and structures within the neck region. In a CT scan, different tissues appear in varying shades of gray, with bone appearing white, air appearing black, and soft tissues appearing in various shades of gray.\\

In this CT scan, the structures appear to be normal. The soft tissues around the neck, including the muscles and fat, appear to be in their expected positions and densities. There are no obvious signs of abnormal masses, calcifications, or other anomalies that would suggest a pathological condition.\\

Therefore, based on the visual inspection of the CT scan, there does not appear to be any perceptible anomaly present. \\

\texttt{\textless/think\textgreater} \\
\texttt{\textless answer\textgreater A\textless/answer\textgreater}
\quad \\
\textbf{Evo-PI's answer (evolved)}:

\texttt{\textless think\textgreater}
1. The image provided is a CT scan of the neck, which is typically used to visualize soft tissues and structures within the neck.\\
\quad \\

2. The scan shows a cross-sectional view of the neck, including the thyroid gland, which is a critical structure in the neck.\\
\quad \\

3. The thyroid gland appears to be in its normal position, with no visible signs of enlargement or abnormal shape.\\
\quad \\

4. There are no visible masses, calcifications, or other abnormalities that would suggest a pathological condition.\\
\quad \\

5. The surrounding structures, such as the trachea and carotid arteries, appear to be in their normal anatomical positions.\\
\quad \\

6. The image does not show any signs of inflammation, infection, or other conditions that would typically be indicated by abnormal findings on a CT scan.\\
\quad \\

Based on the above observations, there does not appear to be any perceptible anomaly present in the image.\\

\texttt{\textless/think\textgreater} \\
\texttt{\textless answer\textgreater A\textless/answer\textgreater}

\end{tcolorbox}

\begin{tcolorbox}[title=Sample of MI case, breakable]
\centering
\includegraphics[width=0.2\linewidth]{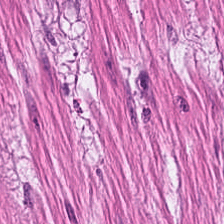} \\[6pt]

\raggedright 

\textbf{Problem:} What is the name of the tissue shown in the picture? \\
\quad A)Nervous tissue \quad B)Cardiac muscle \quad C)Bone tissue \quad D)Smooth muscle

\textbf{Ground Truth:} D \\
\quad \\
\textbf{Huatuo-vision's answer}:

\texttt{\textless think\textgreater}
TheThe image shows a histological section with elongated cells arranged in parallel bundles, which is characteristic of cardiac muscle tissue. The cells have striations and are organized in a way that suggests they are part of the heart's muscular structure.
\texttt{\textless/think\textgreater}
\texttt{\textless answer\textgreater B\textless/answer\textgreater}

\quad \\
\textbf{Evo-PI's answer:}:

\texttt{\textless think\textgreater}
The image shows a microscopic view of tissue with distinct characteristics. The tissue appears to have elongated, striated fibers, which are typical of muscle tissue. The fibers are arranged in parallel bundles, and there are visible nuclei within the fibers, which is a common feature of muscle cells.\\

Let's analyze the options:\\

A) Nervous tissue: Nervous tissue typically consists of neurons and glial cells, which do not have the striated appearance seen in the image.\quad \\

B) Cardiac muscle: Cardiac muscle is found in the heart and has a distinctive striated pattern, but it also has a different arrangement and structure compared to the tissue in the image.\\

C) Bone tissue: Bone tissue is composed of osteocytes embedded in a matrix of bone, which does not match the appearance of the tissue in the image.\\

D) Smooth muscle: Smooth muscle is found in the walls of blood vessels, the digestive tract, and other organs and has a different appearance compared to the tissue in the image.\\

Given the striated appearance and the parallel arrangement of fibers, the tissue in the image is most consistent with smooth muscle.\\

\texttt{\textless/think\textgreater}\\
\texttt{\textless answer\textgreater A\textless/answer\textgreater}

\end{tcolorbox}

\begin{tcolorbox}[title=Sample of MR case, breakable]
\centering
\includegraphics[width=0.2\linewidth]{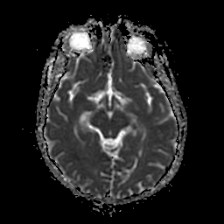} \\[6pt]

\raggedright 

\textbf{Problem:} Is there anything unusual or irregular visible in this image?\\
\quad A)It's difficult to determine if there is an abnormality based on this image. \\
\quad B)There is a possibility of abnormality, but further tests are needed for confirmation. \\
\quad C)I'm not sure, but there might be a slight abnormality. \\
\quad D)No, It's normal.

\textbf{Ground Truth:} A \\
\quad \\
\textbf{Huatuo-vision's answer}:

\texttt{\textless think\textgreater}
The image provided appears to be a brain MRI scan. In MRI scans, variations in signal intensity can indicate different tissue types or abnormalities. The bright areas could represent regions with increased water content, which may suggest edema or inflammation. However, without additional clinical context or comparative images, it is challenging to definitively state whether these findings are abnormal. \\
\quad \\
Given the options:\\
\quad \\
- A) It's difficult to determine if there is an abnormality based on this image.\\
- B) There is a possibility of abnormality, but further tests are needed for confirmation.\\
- C) I'm not sure, but there might be a slight abnormality.\\
- D) No, It's normal.\\
\quad \\
The most appropriate answer would be B) There is a possibility of abnormality, but further tests are needed for confirmation. This is because while the image shows some variations that could be indicative of an abnormality, a definitive diagnosis cannot be made solely from this single image without additional information and possibly further imaging or clinical correlation.\\
\quad \\
\texttt{\textless/think\textgreater}\\
\texttt{\textless answer\textgreater B\textless/answer\textgreater} 

\quad \\

\textbf{Evo-PI's answer}:

\texttt{\textless think\textgreater}
1. Identify the Image Type and Purpose: The image is an MRI (Magnetic Resonance Imaging) scan of the brain, specifically an axial view (horizontal cross-section). This type of imaging is commonly used to visualize the brain's structures and detect abnormalities.\\
\quad \\
2. Analyze the Image Content:\\
- The image shows a clear, well-defined brain structure with no apparent distortions or irregularities.\\
- The white matter and gray matter are distinguishable, and there are no signs of abnormal hyperintensities (bright areas) or hypointensities (dark areas) that would suggest pathology.\\
- The ventricles and sulci (grooves) are visible and appear normal in size and shape.\\
- The skull and surrounding tissues are also clearly visible without any signs of abnormal thickening or thinning.\\
\quad \\
3. Consider the Context and Medical Knowledge:\\
- MRI scans are highly sensitive and specific for detecting \\abnormalities in the brain, including tumors, lesions, and structural anomalies.\\
- The absence of any abnormal signals or changes in the brain's structure in this image suggests that there are no immediate signs of pathology.\\
- The clarity and detail of the image indicate that it is a high-quality scan, which further supports the absence of any abnormalities.\\
\quad \\
4. Evaluate the Options:\\
- Option A: "It's difficult to determine if there is an abnormality based on this image." This option suggests uncertainty, which is not supported by the clear and detailed nature of the image.\\
- Option B: "There is a possibility of abnormality, but further tests are needed for confirmation." This option implies a need for additional testing, which is not indicated by the image alone.\\
- Option C: "I'm not sure, but there might be a slight abnormality." This option suggests uncertainty, which is not supported by the clear and detailed nature of the image.\\
- Option D: "No, It's normal." This option aligns with the clear and detailed nature of the image, indicating that there are no signs of abnormality.\\
\quad \\
5. Conclusion:\\
- The MRI scan of the brain shows no signs of abnormality, as evidenced by the clear and detailed structure of the brain, the absence of any hyperintensities or hypointensities, and the normal appearance of the ventricles and sulci.\\
- The high quality and clarity of the image further support the conclusion that there are no abnormalities present.\\
\quad \\
6. Final Analysis:\\
- The image does not show any signs of abnormality, and the clear and detailed structure of the brain suggests that the scan is normal.\\
- The options provided do not align with the clear and detailed nature of the image, which supports the conclusion that the scan is normal.\\
\quad \\
7. Final Decision:\\
- Based on the detailed analysis and the clear nature of the image, the correct answer is that there is no abnormality present, and the scan is normal.\\
\quad \\
Therefore, the final decision is that the image is normal, and no further tests are needed for confirmation.\\
\quad \\
\texttt{\textless/think\textgreater}
\texttt{\textless answer\textgreater D\textless/answer\textgreater}

\end{tcolorbox}

\end{document}